\documentclass[10pt,twocolumn,letterpaper]{article}

\usepackage{cvpr}              % To produce the CAMERA-READY version
\usepackage{graphicx}
\usepackage{amsmath}
\usepackage{amssymb}
\usepackage{booktabs}
\usepackage{multirow}

\usepackage[pagebackref,breaklinks,colorlinks]{hyperref}

\usepackage[capitalize]{cleveref}
\crefname{section}{Sec.}{Secs.}
\Crefname{section}{Section}{Sections}
\Crefname{table}{Table}{Tables}
\crefname{table}{Tab.}{Tabs.}

\begin{document}

\title{STSM: Spatio-Temporal Shift Module for Efficient Action Recognition}

\author{Zhaoqilin Yang\\
\and
Gaoyun An\\
}
\maketitle

%%%%%%%%% ABSTRACT
\begin{abstract}
   The modeling, computational cost, and accuracy of traditional Spatio-temporal networks are the three most concentrated research topics in video action recognition. The traditional 2D convolution has a low computational cost, but it cannot capture the time relationship; the convolutional neural networks (CNNs) model based on 3D convolution can obtain good performance, but its computational cost is high, and the amount of parameters is large. In this paper, we propose a plug-and-play Spatio-temporal Shift Module (STSM), which is a generic module that is both effective and high-performance. Specifically, after STSM is inserted into other networks, the performance of the network can be improved without increasing the number of calculations and parameters. In particular, when the network is 2D CNNs, our STSM module allows the network to learn efficient Spatio-temporal features. We conducted extensive evaluations of the proposed module, conducted numerous experiments to study its effectiveness in video action recognition, and achieved state-of-the-art results on the kinetics-400 and Something-Something V2 datasets.
\end{abstract}

%%%%%%%%% BODY TEXT
\section{Introduction}
\label{I:0}

With the rapid development of the Internet, camera equipment, and mobile phones, video data has exploded in recent years. The huge amount of video information has far exceeded the processing power of traditional artificial systems, which has aroused people's research interest in video understanding. As a basic task in video understanding, video action recognition has become one of the most active research topics. 
It is widely used in the fields of video surveillance, video retrieval, and social security. With the advancement of science and technology, the requirements for high recognition accuracy and low reasoning complexity are getting higher and higher.

Recently, significant progress has been made in video action recognition based on deep convolutional networks \cite{Simonyan_2014_NIPS,Wang_2016_ECCV,Carreira_2017_CVPR,Lin_2019_ICCV,Li_2020_CVPR,Yang_2020_CVPR,Liu_2021_CVPR,Wang_2021_CVPR_ACT}.
Action recognition CNNs based on 3D convolution, such as C3D \cite{Tran_2015_ICCV} and I3D \cite{Carreira_2017_CVPR}, can most intuitively enable the network to learn Spatio-temporal features. 
However, I3D has learned Spatio-temporal features at the cost of hundreds of GFLOPs and achieved good performance.
Since the use of 3D convolution will cause the network to have a large number of parameters and require more calculations, it greatly limits the practicability of these methods.

\begin{figure}[h]
	\begin{center}
		\includegraphics[width=\linewidth]{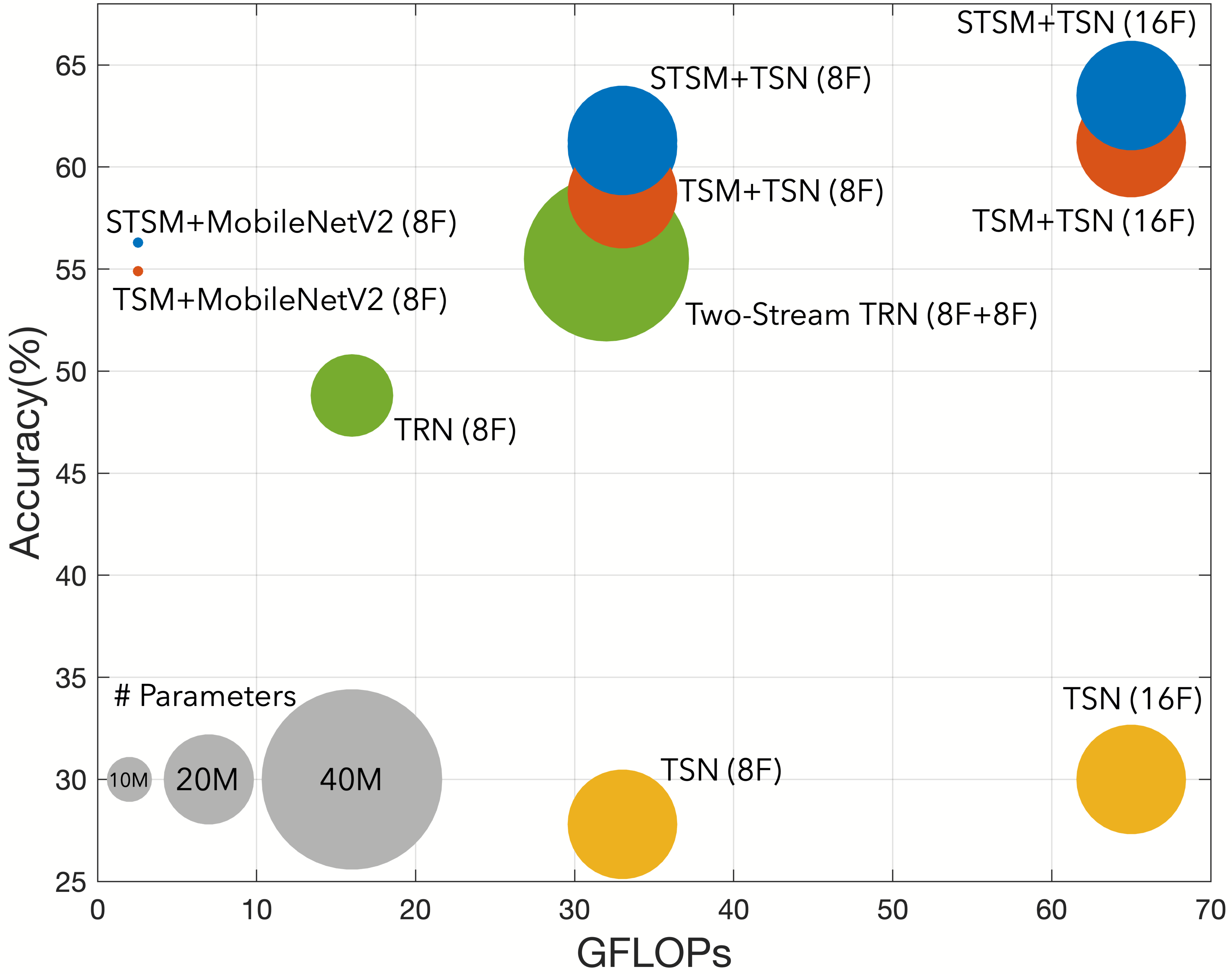}
	\end{center}
	\caption{Our STSM achieves state-of-the-art performance on Something-Something V2 dataset  \cite{Goyal_2017_ICCV} and gets a better accuracy computation trade-off than TSM  \cite{Lin_2019_ICCV}, TRN \cite{Zhou_2018_ECCV}, and TSN \cite{Wang_2016_ECCV}.}
	\label{fig:show}
\end{figure}

Then, in order to reduce the amount of parameters and the amount of calculation, some works \cite{Qiu_2017_ICCV,Xie_2018_ECCV,Tran_2018_CVPR} decompose the 3D convolution kernel into the space part (e.g., $1\times3\times3$) and the time part (e.g., $3\times1\times1$). However, in practice, they add a lot of computational overhead compared with the corresponding 2D convolutions and obtain higher performance with more parameters and computational costs. The recent state-of-the-art model TSM \cite{Lin_2019_ICCV} has achieved a good balance between complexity and performance. It abandons the traditional time convolution and learns time features by moving features along the time dimension by using a Shift operation with zero calculation cost and zero parameter amount. After fusing the convolution results, it is combined with the backbone of 2D CNNs to obtain the most advanced performance with a small amount of computational cost and parameters. This motivates us to focus on designing plug-and-play modules with zero computational cost and zero parameters that can effectively learn Spatio-temporal features.

In this article, we try to design a universal plug-and-play module to make the action recognition model both efficient and high-performance.
First of all, we suggest to tap the potential of Shift convolution, from multiple angles to the dimension $T\times{H}\times{W}$ ($\text{Time}\times\text{Height} \times\text{ Width}$) tensor performs Shift operation to improve network performance.
Therefore, we propose a universal plug-and-play space-time shift module (STSM) with zero calculation cost and zero parameters.
It uses the one-dimensional Shift operation in the T, H, and W dimensions to capture spatiotemporal information from multiple perspectives, and with convolution operations, it can learn spatiotemporal features in one, two, and even higher dimensions.
Taking ResNet as an example, similar to the selection of TSM, we add STSM before the first convolutional layer of each residual block.
The feature tensor is divided by channel, some of the channels perform different Shift operations to learn multi-view features, and the remaining channels remain unchanged.
Then build the final network by embedding STSM in each residual block.
We have conducted extensive experiments on multiple well-known large datasets, including Kinetics-400 \cite{kay_2017_arXiv}, Something-Something V2 \cite{Goyal_2017_ICCV}. The experimental results fully demonstrate the superiority of our STSM.  As shown in Figure \ref{fig:show}, STSM achieves excellent performance with quite limited overhead on Something-Something V2, and it is superior compared with existing state-of-the-art methods. The same conclusion can be drawn on other datasets.

The contributions of our paper are summarized as follows:
\begin{itemize}
	\item We provide a new perspective for efficient video model design by performing Shift operations in different dimensions. It does not require calculations and parameters, but has strong Spatio-temporal modeling capabilities.
	
	\item The STSM module can be easily integrated with existing 2D CNNs or 3D CNNs backbones in a plug-and-play manner. The most advanced performance can be obtained by embedding STSM into other mainstream action recognition models.
	
	\item Compared with existing methods, we have obtained the most state-of-the-art or competitive results in 2 large datasets without increasing the computational cost and the number of parameters.
\end{itemize}
\section{Related work}
\label{Rw:0}

\paragraph{2D CNNs}
2D CNNs are widely used in various fields of deep learning \cite{Krizhevsky_2012_NIPS,Girshick_2014_CVPR,Goodfellow_2014_NIPS,Simonyan_2015_ICLR,Redmon_2016_CVPR}.
Inspired by the great success of deep convolution frameworks in image recognition \cite{Simonyan_2015_ICLR,Ioffe_2015_ICML,He_2016_CVPR}, people initially proposed many methods based on 2D CNNs to apply deep learning to the field of video action recognition.
In these methods, based on the Two-Stream architecture of 2D CNNs \cite{Simonyan_2014_NIPS,Zhang_2016_CVPR} video features can be learned from RGB stream and optical flow stream or motion vectors, respectively, and the output of the two-streams can be fused to obtain the inference result.
TSN \cite{Wang_2016_ECCV} adds a sparse time sampling strategy to the two-stream structure to further improve performance.
TRN \cite{Zhou_2018_ECCV} uses the multi-scale temporal relationship between sampled frames to improve model performance.
Recently, STM \cite{Jiang_2019_ICCV}, GST \cite{Luo_2019_ICCV}, GSM \cite{Sudhakaran_2020_CVPR}, TEINet \cite{Liu_2020_AAAI} focuses on solving the problem of efficient time modeling.
TSM \cite{Lin_2019_ICCV} proposes a universal and effective time shift module to enable 2D CNNs to learn time features.
TEA \cite{Li_2020_CVPR} proposes a time excitation and aggregation module to capture short-term and long-term time evolution.
TDN \cite{Wang_2021_CVPR_TDN} uses the time difference operator to design a module that can capture multi-scale time information to achieve effective action recognition.

\paragraph{3D CNNs  and (2+1)D CNNs variants}
Because 3D CNNs can learn good Spatio-temporal features, they are widely used in the field of action recognition.
C3D \cite{Tran_2015_ICCV} is the first work to apply 3D CNNs to action recognition, which directly uses 3D convolution to learn the spatiotemporal features of the video.
However, C3D has too many parameters, which makes it more difficult to train than 2D CNNs.
I3D \cite{Carreira_2017_CVPR} initializes the network by inflating the 2D convolution pre-trained by ImageNet to 3D convolution, which improves performance while reducing computation time.
S3D \cite{Xie_2018_ECCV}, P3D \cite{Qiu_2017_ICCV}, R(2+1)D \cite{Tran_2018_CVPR}, and StNet \cite{He_2019_AAAI} are inspired by I3D, and can learn Spatio-temporal features while reducing the amount of calculation of 3D convolution.
These (2+1)D CNNs resolve 3D convolutions into 2D spatial convolutions and 1D temporal convolutions.
ECO \cite{Zolfaghari_2018_ECCV} and ARTNet \cite{Wang_2018_CVPR_b} combine 2D and 3D information in CNN blocks to enhance the network's ability to learn features.
Recently, SlowFast \cite{Feichtenhofer_2019_ICCV} explored the use of two different 3D CNN architectures to learn apparent features and motion features, and the two streams are deeply fused to obtain better results.
TPN \cite{Yang_2020_CVPR} proposes a plug-and-play universal time pyramid network at the feature level, which can be flexibly integrated into a 2D or 3D backbone network.
ATFR \cite{Fayyaz_2021_CVPR} improves the energy efficiency of the network by introducing a differentiable Similarity Guided Sampling (SGS) module that can be inserted into any existing 3D CNN architecture.
ACTION-Net \cite{Wang_2021_CVPR_ACT} designed a universal and effective module with 3D convolution and embedded it in 2D CNNs to enable the network to learn spatiotemporal features.
SELFYNet \cite{Kwon_2021_ICCV} is based on the rich and robust motion representation method of spatiotemporal self-similarity to effectively capture the long-term interaction and fast motion in the video, thereby achieving robust motion recognition.

\section{Spatio-temporal Shift Module (STSM)}
\label{STSM:0}

In this section, we first introduce the technical details of our novel Spatio-temporal shift module (STSM) and how to insert it into the ready-made architecture of 2D CNN or 3D CNN. STSM is a plug-and-play module with zero calculation and zero parameters, which can effectively and efficiently encode Spatio-temporal features after being embedded in the target network. Then we describe the relationship between the Shift operation in the STSM module and the convolution operation of the sparse convolution kernel.

\subsection{The Design Spatio-temporal Shift Module}
\begin{figure}[h]
	\begin{center}
		\includegraphics[width=0.8\linewidth]{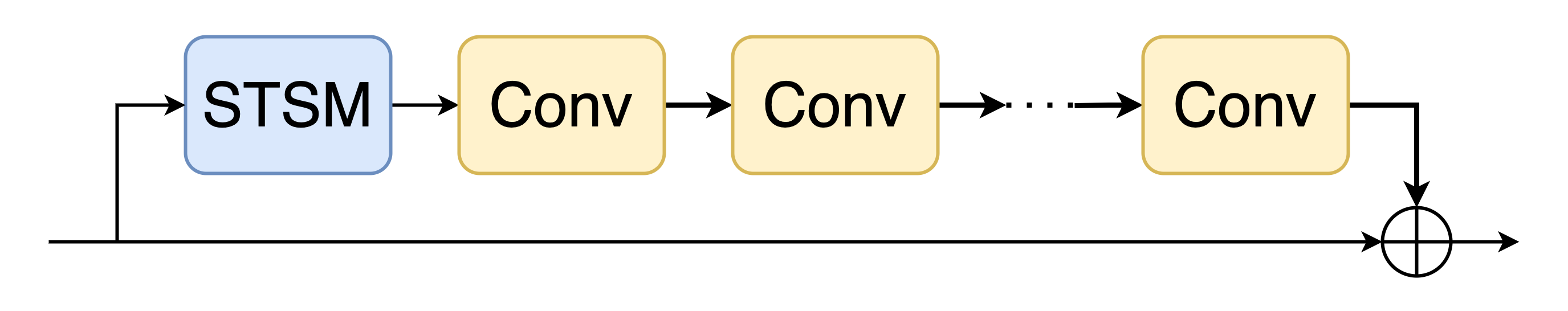}
	\end{center}
	\caption{Insert the STSM module into the residual block. The convolution can be either 2D convolution or 3D convolution.}
	\label{fig:STSM_Res}
\end{figure}

Our STSM module is a plug-and-play module, so as long as it is inserted into any convolutional layer of the network, the network performance can be improved without increasing the number of parameters and calculations. 
For example, for a ResNet structured network, referring to the way that TSM \cite{Lin_2019_ICCV} embeds the time shift module, we insert the STSM module in the way shown in Figure \ref{fig:STSM_Res}. At this time, only one STSM module is inserted for each residual block.  

\begin{figure*}[h]
	\begin{center}
		\includegraphics[width=0.7\linewidth]{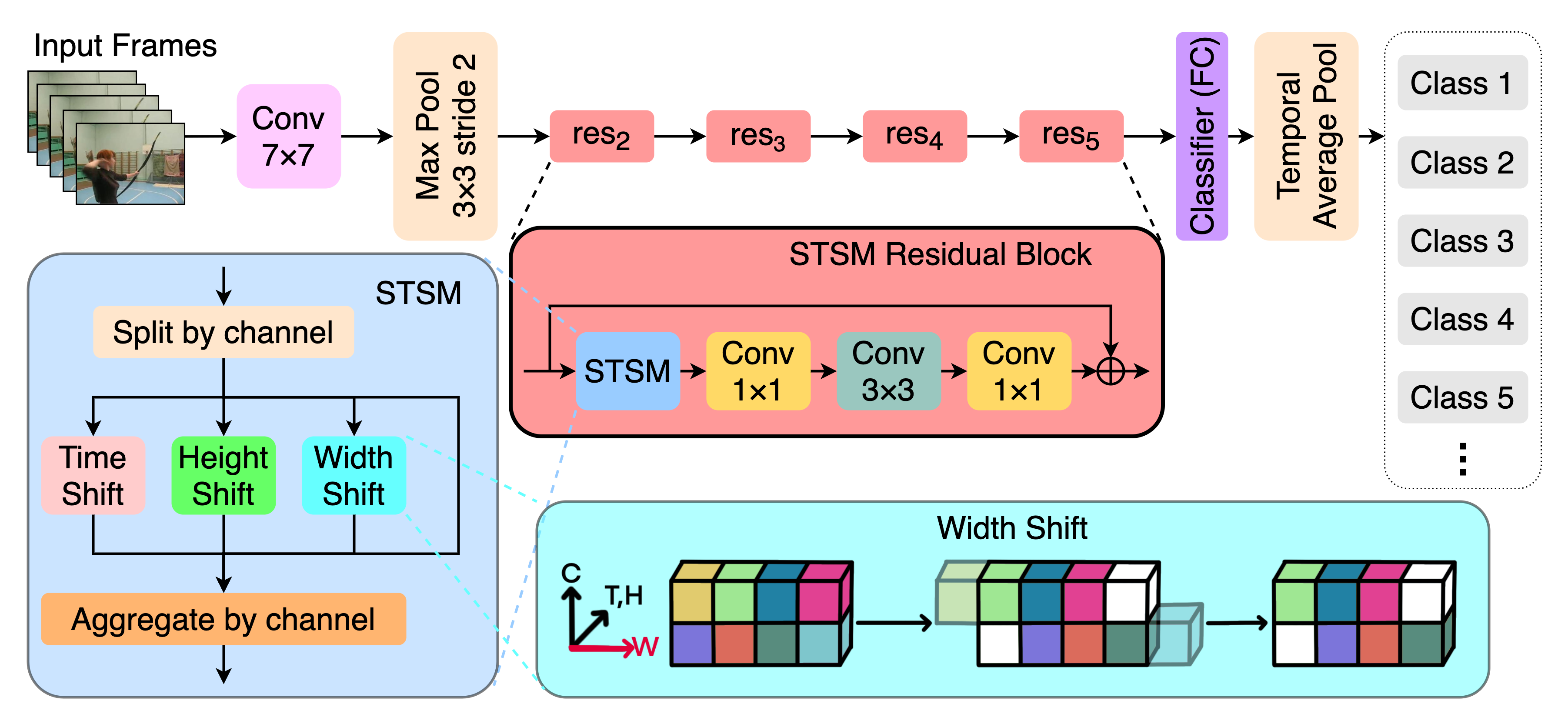}
	\end{center}
	\caption{The network structure after STSM is embedded in ResNet-50 is shown in the figure.
		The convolutions are all two-dimensional spatial convolutions, the first pooling layer is spatial max pooling, and the last pooling layer is an average time pooling. For simplicity, we did not draw the batch normalization layer and the activation function Relu in the figure. }
	\label{fig:STSM_pipeline}
\end{figure*}

Figure \ref{fig:STSM_pipeline} shows the network structure after embedding our STSM module in ResNet-50 \cite{He_2016_CVPR}. 
Because the Shift module is followed by a convolution operation, the result of a two-dimensional Shift operation or even a three-dimensional Shift operation can be obtained through the one-dimensional Shift operation time dimension (T) + height dimension (H) + width dimension (W) and the subsequent convolution operation.

In order to enable the network to adaptively select one-dimensional, two-dimensional, and three-dimensional Shift operations based on samples, our STSM only uses one-dimensional Shift operations based on time dimension + height dimension + width dimension (T+H+W).
Our STSM can learn spatiotemporal features through one-dimensional Shift operation, which is equivalent to that we only use one-dimensional Shift operation T+H+W to achieve adaptive network selection of one-dimensional Shift operation T, H, W, and two-dimensional Shift operation TH, TW, HW, and three-dimensional Shift operate THW to learn Spatio-temporal characteristics. However, TSM only uses the one-dimensional Shift operation in the time dimension, and the network can only learn time features and spatial features separately.

\subsection{What is Spatio-temporal Shift operation?}
\begin{figure}[h]
	\begin{center}
		\includegraphics[width=\linewidth]{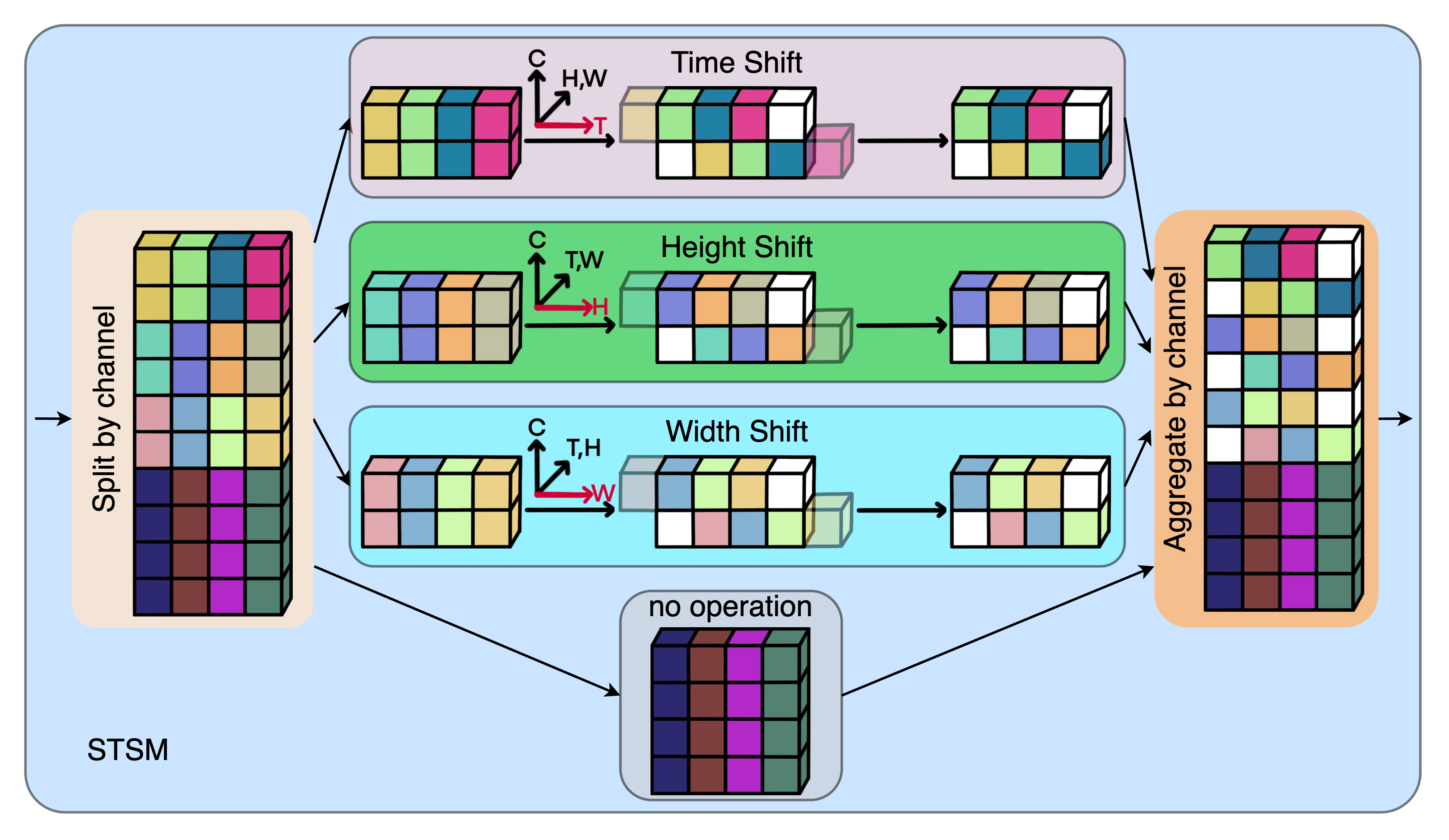}
	\end{center}
	\caption{Time (T) + height (H) + width (W) one-dimensional Spatio-temporal Shift operation schematic diagram.}
	\label{fig:T+H+W}
\end{figure}

Although the principle behind the proposed module is simple, we find that only applying the shift operation to the spatial dimension \cite{Wu_2018_CVPR} or the time dimension \cite{Lin_2019_ICCV} does not fully realize the potential of the Shift operation.
Our proposed STSM module using one-dimensional space-time shift operation is shown in Figure \ref{fig:T+H+W}.
First, the feature tensor is divided into four parts according to the channel.
For the first feature tensor after segmentation, it is divided into two parts according to the channel, and one position is moved forward and backward respectively along the time dimension.
For the second and third feature tensors after segmentation, the same shift operation as the first feature tensor is performed in the height dimension and the width dimension, respectively.
The remaining feature tensors remain unchanged. Finally, we splice the above four feature tensors along the channel dimension to complete our one-dimensional Spatio-temporal Shift operation.

\section{Experimental setting}
\label{Es:0}
This section introduces the experimental settings related to this paper.

\subsection{Datasets and Evaluation Metrics}
We evaluate our method on two large-scale datasets, which are Kinetics-400 (K400) \cite{kay_2017_arXiv}, Something-Something V2 (Sth-V2) \cite{Goyal_2017_ICCV}.
Kinetics-400 has 400 human action categories, and the number of videos is approximately 240k training samples and 20k validation samples.
For the Something-Something V2 dataset, the actions in it have a strong temporal relationship, so it is difficult to classify. It contains 220k videos, and the number of categories is 174 fine-grained categories.

We report the Top-1 accuracy (\%) of Kinetics-400 and Something-Something V2.
In addition, we use computational cost (in FLOPs or GFLOPs) and the number of model parameters to describe model complexity. 
If there are no special instructions, all use ImageNet for pre-training.
\#F and \#Para indicate the number of frames and the number of parameters, respectively.

\subsection{Implementation Details}
Unless otherwise stated, all experiments were performed on MMAction2 \cite{2020mmaction2} using RGB frames and evaluated on the validation set.

\paragraph{Training}
The parameters for training on the Kinetics-400 are: 100 training epochs, initial learning rate $7.5\times10^{-3}$ (decays at epochs 40 and 80 by 0.1), batch size 48, and dropout 0.5.
The entire network uses stochastic gradient descent (SGD) for end-to-end training, with a momentum of 0.9 and a weight decay of $1\times10^{-4}$.
The sample input strategy uses the built-in DenseSampleFrames type of MMAction2, where clip\_len=1, frame\_interval=1, num\_clips=8.
Aiming at the feature that the number of frames of a single video sample in the Something-Something V2 dataset is small. Therefore, the sample input strategy uses the built-in SampleFrames type of MMAction2, where clip\_len=1, frame\_interval=1, and num\_clips=8. The parameters for training on the Something-Something V2 are: 50 training epochs, initial learning rate $7.5\times10^{-3}$ (decays at epochs 20 and 40 by 0.1), batch size 48, and dropout 0.5.
The shortest side of the input frame size will be pre-adjusted to 256 pixels, and then one of the ten-crops will be randomly used to crop the frame to $224\times224$. The ten-crops are top left, top right, center, bottom left, bottom right, and their horizontal mirroring flips.

\paragraph{Inference}
The frame sampling setting during inference is the same as during training. 
After cropping the shortest side of the frame to 256 pixels, then uniformly cropping them into $256\times256$ pixel frames, and finally averaging the output of three-crops to obtain the final output.

\section{Experimental results}
\label{Er:0}

\subsection{Ablation Analysis}
\label{Er:1}

\paragraph{Paramter Choice}
When doing the shift operation, the feature tensor of the current layer will be divided into many parts according to the channel. 
We define the ratio of the number of channels of the shift operation to the total number of channels of the feature tensor as $\alpha$.
For example, when $\alpha=1/4$, if the total number of channels at this time is 64, the Shift operation is performed on the 1st to 16th channels, and the remaining 17th to 64th channels remain unchanged.
Table \ref{tab:1} compares the performance of STSM under different $\alpha$.
We use the 2D ResNet-50 pre-trained by ImageNet as the backbone and embed our STSM in it, and the Shift dimension of STSM is set to T+H+W.
The data in the table is the Top-1 accuracy rate ($\%$) on the validation set of the Kinetics-400 dataset.
%\begin{table}[h]
%	\centering
%	\begin{tabular}{ccc}
%		\hline
%		Setting& Kinetics-400 & Sth-V2  \\
%		\hline
%		$\alpha=1$ &  &  \\
%		$\alpha=3/4$ & 74.49 & 61.31 \\
%		$\alpha=1/2$ & 74.81 &\\
%		$\alpha=3/8$ & 75.04 & 60.98 \\
%		$\alpha=1/4$ & 74.77 &  \\
%		$\alpha=1/8$ &  &  \\
%		$\alpha=0$ &  &  \\
%		\hline
%	\end{tabular}
%	\caption{\label{tab:1} Parameter choices of $\alpha$ about Top-1 accuracy. Backbone: R-50.}
%\end{table}
\begin{table*}[h]
	\centering
	\begin{tabular}{cccccccc}
		\hline
		\multirow{2}*{Setting}& \multicolumn{5}{c}{$\alpha$}  \\
		&0& 1/8& 1/4 & 3/8 & 1/2 &3/4&1\\
		\hline
		Accuracy &72.16&74.62&74.77&\textbf{75.04}&74.81&74.49&73.83  \\
		\hline
	\end{tabular}
	\caption{\label{tab:1} Parameter choices of $\alpha$ about Top-1 accuracy. Backbone: R-50. Dataset: the validation set of the Kinetics-400 dataset.}
\end{table*}

From the results in Table \ref{tab:1}, we can see that our STSM will have different performances under different $\alpha$ settings.
It can be observed that the performance reaches the highest point when $\alpha=3/8$.
When $\alpha<3/8$, the accuracy will decrease as $\alpha$ decreases. When $\alpha>3/8$, the accuracy will decrease with the increase of $\alpha$.
This is equivalent to performing a Shift operation on the first $3/8$ of the channels of the feature tensor, and the remaining channels remain unchanged.
Because the Shift dimension of STSM is set to T+H+W at this time, $\alpha$ with a numerator of 3 will appear.
For simplicity, we treat the levels of each dimension as equal.
It is worth noting that our choice of $\alpha$ is different from TSM ($\alpha=1/4$) \cite{Lin_2019_ICCV}.
The $\alpha=1/4$ in TSM is equivalent to setting $\alpha=3/4$ in our STSM.
In the best $\alpha=3/8$ in our STSM experiment, we only perform Shift operations along the time dimension on the first $1/8$ of the channel.
In the subsequent experiments, unless otherwise specified, our $\alpha$ is set to $3/8$.

\paragraph{Different Shift operations}
Our STSM is a Spatio-temporal Shift convolution module, which can be one-dimensional, two-dimensional, or even higher-dimensional Shift transformation.
How to choose the right Shift dimension is an important issue.
Table \ref{tab:2_1} shows the network performance when we only add one-dimensional and two-dimensional Shift convolutions in the spatial dimension.
Table \ref{tab:2_2} shows the network performance when we add one-dimensional and two-dimensional Shift convolutions to the Spatio-temporal dimension.
\begin{table}[h]
	\centering
	\begin{tabular}{ccccc}
		\hline
		\multirow{2}*{Setting}& \multicolumn{4}{c}{Kinetics-400} \\
		& \#F &  FLOPs & \#Para &  Top-1  \\
		\hline
		TSN (R-50)  from \cite{Lin_2019_ICCV}& 8 & 33G & 24.3M &70.6\\
		T(TSM \cite{Lin_2019_ICCV})     & 8 & 33G & 24.3M &74.1\\
		T(MMAction2) & 8 &33G & 24.3M & \textbf{74.43}  \\
		H                 & 8 & 33G & 24.3M  &72.25 \\
		W               & 8 & 33G & 24.3M  &72.56 \\
		H+W           & 8 & 33G & 24.3M & 72.53\\
		HW            & 8 & 33G & 24.3M  & 72.36\\
		\hline
	\end{tabular}
	\caption{\label{tab:2_1} Network performance under different Shift operations in spatial dimensions. Backbone: R-50. The $\alpha$=1/4, which is consistent with TSM.}
\end{table}
\begin{table}[h]
	\centering
	\begin{tabular}{ccccc}
		\hline
		\multirow{2}*{Setting}& \multicolumn{4}{c}{Kinetics-400} \\
		& \#F &  FLOPs & \#Para &  Top-1  \\
		\hline
		TSN (R-50)  from \cite{Lin_2019_ICCV}& 8 & 33G & 24.3M &  70.6  \\
		T(TSM \cite{Lin_2019_ICCV}) & 8 &33G & 24.3M & 74.1  \\
		T(MMAction2) & 8 &33G & 24.3M & 74.43  \\
		T+H+W & 8 &33G & 24.3M &  \textbf{75.04}  \\
		T+HW & 8 &33G & 24.3M &  74.68 \\
		T+H+W+HW & 8 & 33G & 24.3M & 74.5  \\
		TH+TW+HW & 8 & 33G & 24.3M & 74.95  \\
		T+H+W+TH+TW+HW & 8 & 33G & 24.3M &74.84   \\
		\hline
	\end{tabular}
	\caption{\label{tab:2_2}  In our method, the $\alpha$=3/8. Our method uses the checkpoint of TSM on Kinetics-400 in MMAction2 to initialize the network. Backbone: R-50.}
\end{table}

\begin{table*}[h]
	\centering
	\begin{tabular}{cccccccc}
		\hline
		Model& Backbone & Stream  &Pretrain &\#F &  GFLOPs & \#Para  &   K-400 \\
		\hline
		I3D \cite{Carreira_2017_CVPR}&3D BNInception&RGB&ImageNet& 64&108$\times$N/A& 12M & 71.1 \\
		Two-Stream I3D \cite{Carreira_2017_CVPR}&3D BNInception&RGB+Flow&ImageNet& 64& 216$\times$N/A& 24M  & 74.2 \\
		R(2+1)D \cite{Tran_2018_CVPR}                            &(2+1)D R-50    &RGB&none&16&152$\times$115& 63.6M &72.0\\
		Two-Stream R(2+1)D \cite{Tran_2018_CVPR}                           &(2+1)D R-50    &RGB+Flow&none&16&304$\times$115& 127.2M &73.9\\	
		SlowOnly \cite{Feichtenhofer_2019_ICCV}& 3D R-50 &RGB&ImageNet&4&27.3$\times$3$\times$10& -  &  72.6 \\
		SlowFast \cite{Feichtenhofer_2019_ICCV}&3D R-50&RGB&ImageNet&4$\times$16& 36.1$\times$3$\times$10 & 34.4M  &  75.6\\
		SmallBigNet \cite{Li_2020_CVPR_SmallBigNet}&3D R-50&RGB&ImageNet&8& 57$\times$3$\times$10 & -  & 76.3\\
		TSN-50 + TPN \cite{Yang_2020_CVPR} &3D R-50&RGB&ImageNet&8& - &  - & 73.5\\
		SlowFast+ATFR \cite{Fayyaz_2021_CVPR}&3D R-50&RGB&ImageNet&4$\times$16& 20.8$\times$3$\times$10 & 34.4M  &  75.8\\
		\hline
		TSN from \cite{Lin_2019_ICCV} &2D R-50&RGB&ImageNet& 8 &  33$\times$3$\times$10&24.3M &   70.6\\
		TSN+Mb\_V2 from \cite{Lin_2019_ICCV}&2D Mb\_V2&RGB&ImageNet& 8 &  2.55$\times$3$\times$10&2.33M &   66.5\\
		TSM \cite{Lin_2019_ICCV}&2D R-50&RGB&ImageNet&8 &33$\times$3$\times$10&24.3M&   74.1\\
		TSM+NL \cite{Lin_2019_ICCV}&2D R-50&RGB&ImageNet&8 &-&-&   75.7\\
		TSM+Mb\_V2 \cite{Lin_2019_ICCV}&2D Mb\_V2&RGB&ImageNet&8 &2.55$\times$3$\times$10&2.33M&   69.5\\
		TEA \cite{Li_2020_CVPR}&2D R-50&RGB&ImageNet&8& 35$\times$3$\times$10 & -  & 75.0 \\
		TDN \cite{Wang_2021_CVPR_TDN}  & 2D R-50 & RGB &ImageNet& 8  & 36$\times$3$\times$10 & -& 76.6 \\
		\hline
		STSM+TSN&2D R-50&RGB&ImageNet& 8  &  33$\times$3$\times$10&24.3M &  75.0 \\
		STSM+TSN+NL&2D R-50&RGB&ImageNet& 8  &  -&- &  75.9 \\
		STSM+Mb\_V2 &2D Mb\_V2&RGB&ImageNet& 8  &  2.55$\times$3$\times$10&2.33M &  69.9 \\
		\hline
	\end{tabular}
	\caption{\label{tab:4} The comparison between our method and the state-of-the-art method on the validation set of Kinetics-400 dataset.}
\end{table*}
\begin{table*}[h]
	\centering
	\begin{tabular}{cccccccc}
		\hline
		Model& Backbone & Stream  &Pretrain &\#F &  GFLOPs & \#Para & Sth-V2  \\
		\hline
		MultiScale TRN \cite{Zhou_ECCV_2018} & 3D BNInception &RGB&ImageNet&8& 16$\times$3$\times$2  & 18.3M & 48.8 \\
		Two-Stream TRN \cite{Zhou_ECCV_2018} & 3D BNInception &RGB+Flow&ImageNet& 8+8& 32 $\times$3$\times$2& 36.6M & 55.5 \\
		SmallBigNet \cite{Li_2020_CVPR_SmallBigNet}&3D R-50&RGB&ImageNet&8& 52$\times$3$\times$2 & -  & 59.7 \\
		SmallBigNet \cite{Li_2020_CVPR_SmallBigNet}&3D R-50&RGB&ImageNet&16& 105$\times$3$\times$2 &  - & 62.3 \\
		TSN-50 + TPN \cite{Yang_2020_CVPR} &3D R-50&RGB&ImageNet&8& - &  - & 55.2\\
		SlowFast from  \cite{Fayyaz_2021_CVPR}&3D R-50&RGB&Kinetics-400&4$\times$16& 132.8 & 34.4M  & 61.7 \\
		SlowFast+ATFR  \cite{Fayyaz_2021_CVPR}&3D R-50&RGB&Kinetics-400&4$\times$16& 87.8 & 34.4M  & 61.8 \\
		ACTION-Net \cite{Wang_2021_CVPR_ACT} &3D R-50 &RGB&ImageNet&8  & 34.75$\times$3$\times$2 & 28.08M& 62.5 \\
		ACTION-Net \cite{Wang_2021_CVPR_ACT} &3D R-50&RGB&ImageNet& 16  &  - & 28.08M & 64\\
		ACTION-Net+Mb\_V2\cite{Wang_2021_CVPR_ACT} &3D Mb\_V2&RGB&ImageNet& 8  & 2.57$\times$3$\times$2 & 2.36M& 58.5 \\
		SELFYNet-R50 \cite{Kwon_2021_ICCV}& 3D R-50 & RGB &ImageNet& 8  & 37$\times$3$\times$2 & -& 62.7 \\
		\hline
		TSN from  \cite{Wang_2021_CVPR_ACT}&2D R-50&RGB&Kinetics-400& 8 &  33$\times$3$\times$2&24.3M &  27.8\\
		TSN from  \cite{Wang_2021_CVPR_ACT} &2D R-50&RGB&Kinetics-400& 16 &  65$\times$3$\times$2&24.3M &  30\\
		TSM from \cite{Wang_2021_CVPR_ACT}&2D R-50&RGB&Kinetics-400&8 &33$\times$3$\times$2&24.3M&  58.7\\
		TSM from \cite{Wang_2021_CVPR_ACT}&2D R-50&RGB&Kinetics-400& 16&65$\times$3$\times$2&24.3M & 61.2\\
		TSM+Mb\_V2 from \cite{Wang_2021_CVPR_ACT}&2D Mb\_V2&RGB&ImageNet&8 &2.55$\times$3$\times$2&2.33M&  54.9\\
		TDN \cite{Wang_2021_CVPR_TDN}  & 2D R-50 & RGB &ImageNet& 8  & 36$\times$3$\times$2 & -& 64.0\\
		MG-TSM \cite{Zhi_2021_ICCV}  & 2D R-50 & RGB &ImageNet& 8  & - & -& 60.1 \\
		MG-TEA \cite{Zhi_2021_ICCV}  & 2D R-50 & RGB &ImageNet& 8  & - & -& 62.5\\
		MG-TEA \cite{Zhi_2021_ICCV}  & 2D R-50 & RGB &ImageNet& 16  & - & -& 63.8 \\
		\hline
		STSM+TSN&2D R-50&RGB&ImageNet& 8  &  33$\times$3$\times$2&24.3M & 61.3 \\
		STSM+TSN &2D R-50&RGB&ImageNet& 16  &  65$\times$3$\times$2&24.3M & 63.5 \\
		STSM+Mb\_V2 &2D Mb\_V2&RGB&ImageNet& 8  &  2.55$\times$3$\times$2&2.33M & 56.3 \\
		STSM+Mb\_V2 &2D Mb\_V2&RGB&ImageNet& 16  & 5.1$\times$3$\times$2&2.33M & 59.2 \\
		\hline
	\end{tabular}
	\caption{\label{tab:5} The comparison between our method and the state-of-the-art method on the validation set of Something-Something V2 dataset.}
\end{table*}
From the table \ref{tab:2_1}, it can be seen that on 2D CNNs, the performance of the network using only spatial Shift is improved compared to the backbone network TSN  \cite{Wang_2016_ECCV}, but it is not as good as the network using only time Shift.
Therefore, it is proved that temporal features are essential for 2D CNNs, and it also proves the effectiveness of spatial Shift.
Among them, the network performance of adding a one-dimensional space Shift operation H and W is slightly higher than that of adding a two-dimensional space Shift operation HW. It is proved that after adding multiple one-dimensional space Shift operations to the network, the result after passing through the convolutional layer is equivalent to the result of adaptively selecting one-dimensional, two-dimensional, or higher-dimensional Shift operations.

It can be seen from Table \ref{tab:2_2} that the performance of 2D CNNs with Spatio-temporal shift is better than that of 2D CNNs with temporal shift only.
The most basic one-dimensional Spatio-temporal Shift combination has the best performance and the simplest Shift operation required. Therefore, our STSM module chooses to use only the combination of Time Shift + Height Shift + Width Shift (T+H+W).
The performance of the one-dimensional Shift module with only T+H+W is slightly better than other Spatio-temporal Shift combinations. It can be seen that the one-dimensional Shift operation of T+H+W combined with the convolution operation can allow the network to adaptively select the required Shift operations of different dimensions. Specifically, the network can change one-dimensional Shift T, H, W, two-dimensional Shift TH, TW, TH, and even three-dimensional Shift THW from the one-dimensional Shift combination of T+H+W.
In subsequent experiments, unless otherwise specified, our STSM only uses the one-dimensional Shift operation, and the dimensions are set to T+H+W.

\subsection{Comparison with State-of-the-arts}
\label{Er:2}
As a universal plug-and-play module with 0 parameters and 0 calculations, STSM significantly improves the 2D baseline. We compared the performance of our STSM model with the state-of-the-art methods on Kinetics-400, Something-Something V2.
\paragraph{Kinetics-400}

Kinetics-400 is the current mainstream and challenging large-scale dataset. 
We compare the performance of STSM and the state-of-the-art method on the validation set of the Kinetics-400 dataset in the table \ref{tab:4}. 
Our STSM improves network performance more effectively than TSM without increasing the cost of calculation and the number of parameters.
The network performance of our STSM embedded in 2D CNNs is also competitive with 3D CNNs.
Especially compared with TPN \cite{Yang_2020_CVPR}, which also uses TSN  \cite{Wang_2016_ECCV} as the backbone, we have $1.5\%$ higher accuracy than TPN  with 3D convolution when only 2D convolution is used.
Compared with other networks based on 2D CNNs, our network can achieve higher performance with a lower amount of calculations and parameters.
Compared with TSM \cite{Lin_2019_ICCV}, which is also based on TSN, our network has a higher accuracy rate.
Compared with the original TSN, the accuracy of TSM with TSN as the backbone has increased by $3.5\%$. 
The accuracy of our STSM with TSN as the backbone is increased by $4.4\%$ compared to the original TSN.
Our STSM improves TSN by $22\%$ higher than TSM.
Our STSM also achieves competitive results compared to TEA \cite{Li_2020_CVPR} and TDN \cite{Wang_2021_CVPR_TDN}.
It should be noted that both TEA and TDN are specifically designed for ResNet  \cite{He_2016_CVPR}, and our STSM can be easily embedded in other architectures, such as MobileNetV2  (Mb\_V2) \cite{Sandler_2018_CVPR} and ResNet-50 + Non-local module (NL) \cite{Wang_2017_CoRR}.
In particular, our STSM is a plug-and-play ultra-lightweight module with zero parameters and zero calculations, while TEA and TDN are both complex modules that increase calculations and parameters.

\paragraph{Something-Something V2}
Something-Something V2 is a challenging large dataset. 
Table \ref{tab:5} shows the performance comparison between our STSM and the state-of-the-art method under the validation set of the Something-Something V2 dataset.
Our STSM obtained more competitive results on this dataset than on Kinetics-400.
First of all, compared with 3D CNNs, our STSM+TSN only has 2D convolution, and when the number of input frames is the same, our network is second only to ACTION-Net  \cite{Wang_2021_CVPR_ACT}. But our STSM+TSN calculation amount and parameter amount are lower than ACTION-Net.
When the comparison methods are all 2D CNNs, our method is more competitive.
When the number of input frames is the same, the result is second only to TDN \cite{Wang_2021_CVPR_TDN} and MG-TEA \cite{Zhi_2021_ICCV}, but TDN increases the amount of calculation, and the reference network of MG-TEA is far superior to TSN \cite{Wang_2016_ECCV}.
The performance of MG-TSM \cite{Zhi_2021_ICCV} with TSM  \cite{Lin_2019_ICCV} as the backbone network is $1.2\%$ lower than ours with TSN as STSM, which proves that our STSM can have a higher degree of improvement than MG \cite{Zhi_2021_ICCV} without increasing the amount of calculation and parameters.

\subsection{Different backbone}
STSM can scale well to backbones of different sizes. 
Table \ref{tab:3_1} and  \ref{tab:3_2} respectively show the results of our STSM after embedding different backbones on the validation sets of Kinetics-400 and Something-Something V2 datasets. 
In the experiment, ResNet-50 (R-50) \cite{He_2016_CVPR}, MobileNetV2 (Mb\_V2) \cite{Sandler_2018_CVPR} and ResNet-50 + Non-local module (NL) \cite{Wang_2017_CoRR} were used as the backbones.
The methods in the two tables have the same GFLOPs and parameters under the same backbone and input frames.
It can be seen from the experimental results that our STSM is effective under different backbones and input frame numbers, and all due to the TSM \cite{Lin_2019_ICCV} under the same backbone and input frame numbers.

\begin{table}[h]
	\centering
	\begin{tabular}{cccc}
		\hline
		\multirow{2}*{Model}& \multicolumn{3}{c}{Kinetics-400} \\
		& Backbone & \#F  &  Top-1\\
		\hline
		TSN from \cite{Lin_2019_ICCV} & R-50 & 8 & 70.6\\
		TSM+TSN \cite{Lin_2019_ICCV} & R-50 & 8 & 74.1 \\
		STSM+TSN & R-50  &  8 & \textbf{75.0}\\
		\hline
		TSN+NL from \cite{Lin_2019_ICCV} & R-50 & 8 & 74.6\\
		TSM+TSN+NL \cite{Lin_2019_ICCV} & R-50 & 8 & 75.7 \\
		STSM+TSN+NL & R-50  &  8 & \textbf{75.9}\\
		\hline
		TSN from \cite{Lin_2019_ICCV} & Mb\_V2& 8  & 66.5\\
		TSM+TSN \cite{Lin_2019_ICCV}& Mb\_V2 & 8  & 69.5\\
		STSM+TSN &  Mb\_V2&8 & \textbf{69.9}\\
		\hline
	\end{tabular}
	\caption{\label{tab:3_1} Our STSM uses different backbone network experiments on the validation set of the Kinetics-400 dataset. In the case of the same backbone, the above methods have the same GFLOPs and parameters.}
\end{table}

\begin{table}[h]
	\centering
	\begin{tabular}{cccc}
		\hline
		\multirow{2}*{Model}& \multicolumn{3}{c}{Something-Something V2} \\
		& Backbone & \#F  &  Top-1\\
		\hline
		TSN from  \cite{Wang_2021_CVPR_ACT} & R-50 & 8 & 27.8\\
		TSM+TSN \cite{Lin_2019_ICCV} & R-50 & 8 & 59.1\\
		TSM+TSN from \cite{Wang_2021_CVPR_ACT} & R-50 & 8 & 57.8\\
		STSM+TSN & R-50  &  8 & \textbf{61.3}\\
		\hline
		TSN from  \cite{Wang_2021_CVPR_ACT} & R-50 & 16 & 30.0\\
		TSM+TSN \cite{Lin_2019_ICCV} & R-50 & 16 & 63.4\\
		TSM+TSN from \cite{Wang_2021_CVPR_ACT} & R-50 & 16 & 61.2\\
		STSM+TSN & R-50  &  16 & \textbf{63.5}\\
		\hline
		TSM+TSN from \cite{Wang_2021_CVPR_ACT}& Mb\_V2 & 8  & 54.9\\
		STSM+TSN &  Mb\_V2&8 & \textbf{56.3}\\
		\hline
	\end{tabular}
	\caption{\label{tab:3_2} Under the different number of input frames and backbone network, our STSM compares the accuracy of the backbone network with the validation set of the Something-Something V2 data set.
	In the case of the same backbone, the above methods have the same GFLOPs and parameters.}
\end{table}

\section{Conclusion}
\label{C:0}
We propose a Spatio-temporal shift module for efficient video recognition. It can be inserted into the network backbone in a plug-and-play manner to enhance network performance without increasing the amount of calculation and parameters, especially to enable 2D CNN to learn Spatio-temporal information. The module moves part of the channels in the time dimension and the space dimension on different channels so that the network can learn the Spatio-temporal characteristics.

%%%%%%%%% REFERENCES
{\small
\bibliographystyle{ieee_fullname}
\bibliography{egbib}
}

\end{document}